\documentclass[runningheads]{llncs}
\usepackage{cite}
\usepackage{amsmath,amssymb,amsfonts,mathtools}
\DeclareMathOperator*{\argmin}{arg\,min}

\usepackage[linesnumbered,lined,ruled]{algorithm2e}
\usepackage{algpseudocode}

\usepackage{graphicx}
\graphicspath{{images}}

\usepackage{textcomp}
\usepackage{xcolor}
\usepackage[caption=false]{subfig}
\usepackage{hyperref}
\usepackage{todonotes}
\usepackage[]{algorithm2e}
\usepackage{capt-of}  % <---
\usepackage{cuted}    % <===
\usepackage{lipsum}
\usepackage{orcidlink}

\usepackage{multirow}
\usepackage{booktabs}
\usepackage{array}
\newcolumntype{X}[1]{>{\centering\let\newline\\\arraybackslash\hspace{0pt}}m{#1}}

\usepackage{xcolor}
\definecolor{codegray}{rgb}{0.5,0.5,0.5}

\newcommand{\reffig}[1]{Fig.~\ref{#1}}
\newcommand{\reftab}[1]{Table~\ref{#1}}
\newcommand{\refsec}[1]{Sec.~\ref{#1}}

\newcommand{\refalg}[1]{Algorithm~\ref{#1}}

\begin{document}

\title{
    Planning Robot Placement for Object Grasping
    \thanks{This work was supported by the Bonn-Aachen International Center for Information Technology (b-it).}
}
%
% {\footnotesize \textsuperscript{*}Note: Sub-titles are not captured in Xplore and
% should not be used}
% \thanks{Identify applicable funding agency here. If none, delete this.}
% }
\author{
    Manish Saini\inst{1} \and
    Melvin Paul Jacob\inst{1} \and
    Minh Nguyen\inst{2}\orcidlink{0000-0002-0811-6441} \and
    Nico Hochgeschwender\inst{2}\orcidlink{0000-0003-1306-7880}}
\authorrunning{M. Saini et al.}
% First names are abbreviated in the running head.
% If there are more than two authors, 'et al.' is used.
%
\institute{
Hochschule Bonn-Rhein-Sieg, Germany
\email{\{manish.saini,melvin.paul\}@smail.inf.h-brs.de}
\and
University of Bremen, Germany
\email{\{minh.nguyen,nico.hochgeschwender\}@uni-bremen.de}
}

% \author{\IEEEauthorblockN{Manish Saini$^{\ast}$}
% ~\\
% \and
% \IEEEauthorblockN{Melvin Paul Jacob$^{\ast}$}
% ~\\
% \and
% \IEEEauthorblockN{Minh Nguyen$^{\dagger}$~\orcidlink{0000-0002-0811-6441}
% ~\\
% \and
% Nico Hochgeschwender$^{\dagger}$~\orcidlink{0000-0003-1306-7880}
% }
% \IEEEauthorblockA{\textit{Hochschule Bonn-Rhein-Sieg} \\
% Sankt Augustin, Germany \\
% minh.nguyen@h-brs.de}
% ~\\
% \and
% \IEEEauthorblockN{Nico Hochgeschwender$^{\dagger}$~\orcidlink{0000-0003-1306-7880}}
% ~\\
% \thanks{$^{\ast}$~Dept. of Computer Science, Hochschule Bonn-Rhein-Sieg, Germany. %
% \texttt{\{manish.saini,melvin.paul\}@smail.inf.h-brs.de}} %
% \thanks{$^{\dagger}$~Faculty of Math. \& Computer Science, University of Bremen, Germany. %
% \texttt{\{minh.nguyen,nico.hochgeschwender\}@uni-bremen.de}} %
% \and
% \IEEEauthorblockN{6\textsuperscript{th} Given Name Surname}
% \IEEEauthorblockA{\textit{dept. name of organization (of Aff.)} \\
% \textit{name of organization (of Aff.)}\\
% City, Country \\
% email address or ORCID}
% }

\maketitle

\begin{abstract}
% \todo[inline]{Highlight the message: prominent approaches typically relies on costly grasp planner to generate grasp candidates, and evaluate possible base placement for these candidates. Here, we propose the idea of finding base placements that make the grasping easier first, then initiating the manipulation behvaiour.}
When performing manipulation-based activities such as picking objects, a mobile robot needs to position its base at a location that supports successful execution.
% A typical solution to this problem involves a robot moving to a predefined location near the location of the target object and then initiating the manipulation behaviour.
To address this problem, prominent approaches typically rely on costly grasp planners to provide grasp poses for a target object, which are then are then analysed to identify the best robot placements for achieving each grasp pose.
% \todo[inline]{highlight base placement before grasp planning, not the other way around, stress service robotics require mobile manipulation}
% However, this approach does not always ensure successful execution, as many factors such as the robot's physical constraints and environmental obstruction may hinder a successful grasp.
In this paper, we propose instead to first find robot placements that would not result in collision with the environment and from where picking up the object is feasible, then evaluate them to find the best placement candidate.
Our approach takes into account the robot's reachability, as well as RGB-D images and occupancy grid maps of the environment for identifying suitable robot poses.
The proposed algorithm is embedded in a service robotic workflow, in which a person points to select the target object for grasping.
We evaluate our approach with a series of grasping experiments, against an existing baseline implementation that sends the robot to a fixed navigation goal.
The experimental results show how the approach allows the robot to grasp the target object from locations that are very challenging to the baseline implementation.
\end{abstract}
%we should not talk about the arrangement of the objects as we are not showing that in the experiments.

% \begin{IEEEkeywords}
% Navigation, Planning, Grasping, Manipulation
% \end{IEEEkeywords}

\section{Introduction}\label{sec:intro}
A common task for mobile robots in domestic settings is object retrieval, e.g. fetching a cup from a dining table or kitchen counter.
Solutions for this task typically involve the robot navigating to a predetermined location and then initiating the picking-up behaviour.
% Typically, these planners determine a location close to the object that satisfies some minimal criteria without taking into account the success rate of grasping the desired object from that location.  % Our approach also doesn't consider the success rate
Such approaches often do not take into account factors of the environment or physical limitations of the robot, which can result in the robot trying to grasp the object from awkward positions, for instance, a far table edge rather than one that is close to the object.
Such suboptimal placements can increase the difficulty of the manipulation task, increasing the risk of failure or even making the task infeasible, e.g. when the target object cannot be reached without the robot colliding with the table or with other objects.
% The robot may also position itself where it encounters obstacles between itself and the target object, complicating the retrieval process. % Since we didn't do these experiments, I'd leave this out
% Such suboptimal base placements can lead to inefficiencies, necessitating additional travel distances and possibly causing accidents with obstructions, complicating the object's recovery process.  % Fixed nav goal would not result in additional distance, I'd also refrain from anything recovery-related, as we did not touch that
% Furthermore, the assumption of fixed navigation goals limits the robot's capacity to adapt to the changing configurations of things on the table, limiting its ability to traverse smoothly.  % I don't think we did anything towards "smoothness", I'd also be careful with "adaptive", as it gives the impression that we change the object configuration at runtime
% Because of the set navigation target, inefficient manipulator motions may occur, increasing the risk of failure in grabbing potential objects and/or robot damage.  % We didn't do anything towards manipulator motion either

% To overcome these obstacles, a more flexible and context-aware navigation technique that takes into account the dynamic character of open spaces in residential situations is required.  % not 'required', we also don't deal with anything 'dynamic'
To overcome these challenges, we propose in this paper a robot placement planning approach that identify locations from where a mobile robot can grasp an object more easily.
Our approach takes into account occupancy grid maps and RGB-D data to check whether the placement candidates can be navigated to or may be at risk of collision during the manipulation motion.
The candidates are then evaluated using a criterion that characterizes the difficulty of the overall navigation and manipulation motion, and the best candidate is selected for carrying out the pickup behaviour.
The planning algorithm is embedded in a service robotics workflow, in which a human operator chooses the target object using hand gestures.
We evaluate our approach with a series of grasping experiments using the Toyota Human Support Robot (HSR) and compare it against a baseline implementation which grasps from a predefined navigation goal.

The remainder of this article is organized as follows.
Related studies will be discussed in the next section.
\refsec{sec:approach} describes in detail our robot placement planning algorithm and how we carry out the pickup behaviour using the generated placement candidates.
\refsec{sec:eval} describes our experimental setup and evaluates the results of our grasping experiments.
Finally, \refsec{sec:conclusion} includes concluding remarks and discusses possible extensions to our approach in the future.
% We control the locations of objects on the table to investigate how our approach can improve grasping performance for object locations deemed difficult for the predefined base location.

\section{Related Work} \label{sec:related}

The traditional approach to improve object grasping performance is grasp planning, which aims to find placement and configuration of end effectors that can best satisfy some criteria relevant to the grasping task~\cite{sahbani2012,bohg2014,kleeberger2020}.
Grasp planning can be broadly categorized into analytical~\cite{sahbani2012} and data-driven~\cite{bohg2014,kleeberger2020} methods, with the former typically analysing the target object's shape and/or hand-object contact to evaluate different properties of grasp candidates, such as stability or task compatibility.
Empirical or data-driven methods~\cite{bohg2014,kleeberger2020}, on another hand, sample and evaluate grasp candidates based on some existing grasp experience, either characterized by a heuristic or generated in simulation or on a real robot.
%Lou et al.~\cite{lou2021} proposes a collision-aware reachability predictor for grasping systems with 6-DoF.
%Using deep neural networks trained through simulation, the system predicts collision-free probability for grasp poses. % I would leave this out for now and opt for the survey citations, as we should probably mention several instead of just one. The wording here also needs improvement and perhaps a closer look into the paper

Another approach for improving the performance of the robot's grasping behaviour is to consider environmental features during motion planning or control of the robot arm~\cite{frank2022}.
This can be done in combination with Dynamic Movement Primitives (DMP), in which complex trajectories, e.g. motion of a robot manipulator, are modelled using a system of second-order linear Ordinary Differential Equation (ODE).
A learnable forcing term is then introduced to adapt these trajectories to some goal-oriented behaviours using well-defined attractor dynamics~\cite{ijspeert2013}.
Data from proximity sensors can then be incorporated into the motion plan, e.g. for obstacle avoidance~\cite{frank2022}, by introducing a coupling term to the DMP formulation.
This coupling term can either be derived analytically via computation of the potential field~\cite{ginesi2019,Tan2011,Hoffmann2009} or learned from sensory data~\cite{sutanto2018}.

%More specifically, Tan et al.~\cite{Tan2011} compute a time-varying target using the potential field and adjust the resulting DMP trajectory, whereas Hoffmann et al.~\cite{Hoffmann2009} add a forcing element to the DMP.
%In \cite{Gams2016}, the human arm has been given a potential field function that was utilized to modify the DMP so that coaching could be performed.
% Ideally also with Probabilistic Movement Primitives

To address the problem of finding an optimal base placement for grasping an object, several approaches utilize inverse reachability maps (IRM) introduced in~\cite{vahrenkamp2013}, which invert a representation of the robot's workspace using its kinematics for a given grasp candidate.
The resulting robot placements are then evaluated using some quality metric, typically to minimize the chance of self-collision or singularity.
Related research has improved upon various aspects of IRM, including quality metrics for possible robot placements~\cite{yao2024} and representations of the robot workspace~\cite{jundong2015,yao2024}.
Burget and Bennewitz~\cite{Burget2015} extend the approach with constraints specific to humanoid robots to find optimal foot placements for a grasp candidate.
Reister et al. \cite{Reister2022} apply IRM to the problem of finding time-efficient base placements for picking and placing several objects consecutively, where the placements are evaluated by a combination of manipulation and navigation costs, and reachability inversion is used for the manipulation portion of the cost function.
Paus et al.~\cite{paus2017} combine IRM with coverage path planning to minimize the number of base repositioning needed to achieve optimal coverage with the manipulator.
IRM is also used as priors in machine learning approaches, e.g. for learning approximations of the robot's inverse reachability~\cite{Welschehold2018} or training motion models~\cite{jauhri2022}.
Aside from IRM-based approaches, Yamazaki et al. \cite{Yamazaki2022} evaluate different ``motions'' for a target end-effector pose by analysing the error distributions associated with carrying out the motion, which consists of moving the base along a trajectory and the arm to a fixed joint configuration.
These approaches all rely on the candidate grasp pose being available, e.g. from a grasp planner, which differs from our approach of finding base placements before planning the grasp motion.

% The first related section gives the general information about the kind of approaches and then provides information around it. Then it discusses few techniques where they lacking in considering environmental factors. But then there are few techniques given as it is, which does no seem relevant

\section{Approach} \label{sec:approach}

\begin{figure}[ht]
    \centering
    \includegraphics[width=\linewidth]{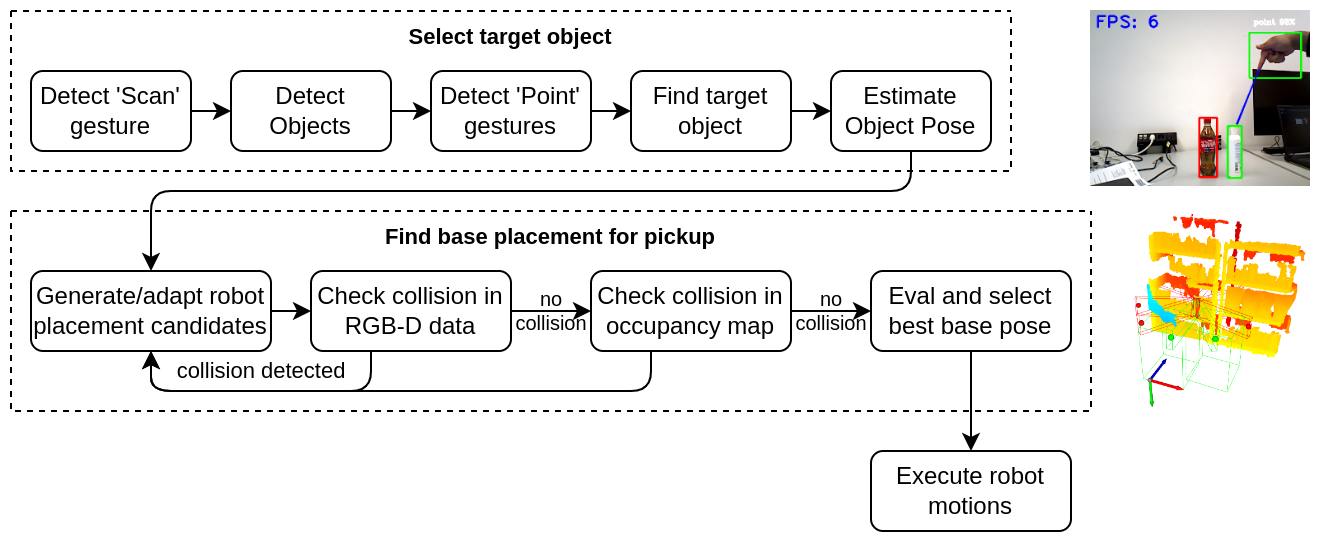}
    \caption{Overall execution flow of the base pose planning component.}
    \label{fig:workflow}
\end{figure}

Our approach to finding base pose placements for picking up objects is embedded in a service robotics workflow, illustrated in \reffig{fig:workflow}).
Here, a human operator chooses the target object using a pointing gesture, then our planning algorithm finds a base pose placement based on the estimated pose of the object, and finally, the according navigation and manipulation motions are planned and executed using existing solutions available on the robot.

\subsection{Selecting the target object using hand gesture}

For gesture detection, we adapt and train an existing MLP architecture\footnote{\url{https://github.com/Kazuhito00/hand-gesture-recognition-using-mediapipe}} to classify pixel values of hand landmarks detected using the MediaPipe library\footnote{\url{https://developers.google.com/mediapipe/solutions/vision/hand_landmarker}} as either \emph{Scan}, \emph{Point}, or \emph{Stop}.
Initially, the component waits for a \emph{Scan} gesture from the operator, after which it detects objects in the scene.
Next, when the \emph{Point} gesture is recognized, a line is fitted to the detected landmarks of the pointing finger and extends from the fingertip until intersecting either the bounding box of a detected object or the image boundary.
The object whose bounding box is intersected is then the target object, and its position in the map frame is estimated from the corresponding depth data in the RGB-D image using the Open3D library~\cite{Zhou2018}.
%, as shown in figure~\ref{fig:gestures:point}.
% The object corresponding to the bounding box where this line intersects for at least 3 seconds will be recognized as the selected object, which is to be retrieved.
% Once we have the bounding box of the selected object we map it to the point cloud to extract the sub-cloud of the selected object. After cleaning the sub-cloud to remove data that does not belong to the object (table, nearby objects, etc.) using clustering and plane fitting, we fit a 3D bounding box to the cleaned sub-cloud cloud to calculate the object pose.
% Processing of the 3D point cloud is done using the Open3D library \cite{Zhou2018}.

\subsection{Planning a robot base placement to pick up the object}

\begin{algorithm}[ht]
\SetAlgoLined
\SetKwInput{KwParam}{Param}
\SetKwFunction{KwCollision}{HasCollisionRisk}
\SetKwFunction{KwBreak}{break}

\KwParam{Radius of robot footprint $r_r$, robot height $h_r$}
\KwParam{Minimum and maximum reach of robot $d_{min}, d_{max}$}
\KwParam{Angle increment for generating radial vectors $\theta$}

\KwIn{Object position in map frame $\prescript{m}{}{\mathbf{p}_o}$}
\KwIn{Detected object width and height $w_o, h_o$}
\KwIn{RGB-D image of the scene $\mathcal{I}$}
\KwIn{Occupancy grid map $\mathcal{G}$}

\KwOut{Set of base placement candidates $\mathcal{S}$}

Set of robot pose candidates: $\mathcal{S} \leftarrow \emptyset$

Generate set of radial unit vectors: $\prescript{m}{}{\mathbf{p}_o}, \theta \mapsto \mathcal{V} $

\ForEach{$\mathbf{\hat{v}}_i \in \mathcal{V}$}{
    Initial robot position candidate w.r.t. object: $\prescript{o}{}{\mathbf{p}_{ri}} \leftarrow d_{min} \mathbf{\hat{v}}_i $

    \While{$ \lVert \prescript{o}{}{\mathbf{p}_{ri}} \rVert < d_{max} $}{
        \If{$\neg$ \KwCollision{$\prescript{o}{}{\mathbf{p}_{ri}}$, $r_r$, $h_r$, $w_o$, $h_o$, $\mathcal{I}$, $\mathcal{G}$}}{
            Pose candidate: $\prescript{o}{}{\mathbf{q}_{ri}} \leftarrow \{\prescript{o}{}{\mathbf{p}_{ri}}, \prescript{o}{}{\mathbf{o}_r}\}$, s.t. robot points in $-\mathbf{\hat{v}}_i$

            \tcc{Transform to map frame and add to set of candidates}
            $\mathcal{S} \leftarrow \mathcal{S} + \{ \prescript{m}{}{\mathbf{q}_{ri}} \}$

            \KwBreak{}
        }

        \tcc{Risk of collision detected, try position further away}
        $\prescript{o}{}{\mathbf{p}_{ri}} \leftarrow (\lVert \prescript{o}{}{\mathbf{p}_{ri}} \rVert + r_r) \mathbf{\hat{v}_i}$
    }
}

\KwRet{$\mathcal{S}$}

\caption{Robot placement planning algorithm.}
\label{alg:base-placement-planning}
\end{algorithm}

Having obtained the object position in the map frame $\prescript{m}{}{\mathbf{p}_o}$, we generate a set of potential base placements from where the pickup behaviour can be executed and rank them using a metric characterizing the difficulty of carrying out the corresponding navigation and manipulation motions.
\refalg{alg:base-placement-planning} shows the pseudocode for our base placement planning component.
For the remainder of this paper, notations of geometric relations will follow the $\prescript{origin frame}{}{relation_{target frame}}$ convention, where poses are denoted with $\mathbf{q}$, positions $\mathbf{p}$, and orientations $\mathbf{o}$.

\subsubsection{Generating robot placement candidates}
First, we project $\prescript{m}{}{\mathbf{p}_o}$ onto the horizontal plane parallel to the ground and generate a set of radial unit vectors $\mathcal{V}$ evenly spaced at an angle $\theta$\footnote{This angle can be calculated based on the robot's reach and footprint radius. For example, the angle to achieve a chord length of $2r_r$ on a circle of radius $d_{min}$ is $\theta = 2 \arcsin{\frac{r_r}{2 d_{min}}}$}, originating from the projected object's position and pointing outward.
Along each $\mathbf{\hat{v}}_i \in \mathcal{V}$, a pose candidate for placing the robot $\prescript{o}{}{\mathbf{q}_{ri}}$ (w.r.t. the object) is selected within a predefined distance range $d_{min} \leq \lVert \prescript{o}{}{\mathbf{p}_{ri}} \rVert \leq d_{max}$ and points towards the object, where the distance range corresponds to the robot's reachable workspace.
This is done by initializing the position candidate to be at distance $d_{min}$ from the object and increasing the distance by $r_r$ until $d_{max}$ if the risk of collision during pickup behaviour is detected.
Repeating this process for each of the radial vectors, we acquire a set of base pose candidates $\mathcal{S}$.

\begin{figure}[ht]
    \centering
    \includegraphics[width=0.65\linewidth]{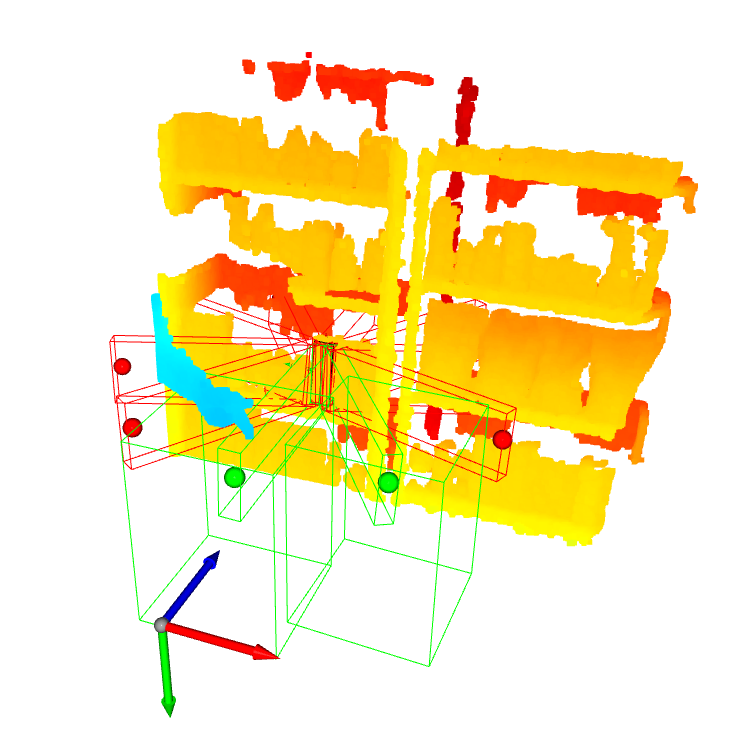}
    \caption{
        Checking robot placement candidates for risk of collision during the pickup behaviour using RGB-D data.
        The regions checked for obstacles are visualized by cuboids.
        Candidates and corresponding cuboids that may result in collision are coloured red, and valid ones green.
    }
    \label{fig:pruning}
\end{figure}

\subsubsection{Checking for risk of collision}
For each position candidate $\prescript{o}{}{\mathbf{p}_{ri}}$, we process the RGB-D image used for object detection to check if there is a risk of collision when the robot reaches the object.
This is done by filtering the point cloud to select only points in the space that the robot may occupy during the pickup behaviour, and eliminate the candidate if the number of points found is higher than a fixed threshold $k_{obs}$.
To this end, we perform the check twice, once for the robot body at the base placement pose and once for the manipulator during the reaching motion.
The checked regions are visualized as the cuboids in \reffig{fig:pruning}, where valid base placement candidates are coloured green, and pruned ones red.
The larger cuboids are of size $2r_r \times 2r_r \times h_r$, corresponding to the robot's dimensions.
Each smaller cuboid extends from the object position to the position candidate at the object's height, and has width and height corresponding to the object's detected width and height $w_o, h_o$.

In addition to RGB-D data, we also process the occupancy grid map provided by the robot to check if placing the robot at $\prescript{o}{}{\mathbf{p}_{ri}}$ may result in a collision.
This is done by transforming a circle of radius $r_r$ centred at $\prescript{o}{}{\mathbf{p}_{ri}}$ to the grid coordinates of the occupancy map and check if the corresponding cells are occupied.
In our experiments, we used only the static occupancy grid map to check for collision, but this step can also be performed on dynamic occupancy maps, if available, to check for dynamic obstacles.

If either of the above checks detects a risk of collision, a new position along $\mathbf{\hat{v}}_i$ at a distance $r_r$ further away from the object is evaluated, and the process repeats until the distance exceeds $d_{max}$.
If no risk of collision is detected, the corresponding pose candidate at position $\prescript{o}{}{\mathbf{p}_{ri}}$ and pointing towards the object, $\prescript{o}{}{\mathbf{q}_{ri}}$, is added to the set of candidates $\mathcal{S}$.

\begin{algorithm}[ht]
\SetAlgoLined
\SetKwInput{KwParam}{Param}
\SetKwFunction{KwMotionCost}{MotionCost}
\SetKwFunction{KwNav}{Navigate}
\SetKwFunction{KwPick}{Pickup}
\SetKwFunction{KwBreak}{break}
\SetKwFunction{KwCont}{continue}

\KwIn{Object position in map frame $\prescript{m}{}{\mathbf{p}_o}$}
\KwIn{Set of base placement candidates $\mathcal{S}$}

\While{$ \mathcal{S} \neq \emptyset $}{
    Read robot current pose $\prescript{m}{}{\mathbf{q}_{r}^c}$
    
    Best candidate: $\prescript{m}{}{\mathbf{q}_{r}^*} \leftarrow \argmin_{\prescript{m}{}{\mathbf{q}_{r}}} $ \KwMotionCost{$\prescript{m}{}{\mathbf{q}_{r}}$, $\prescript{m}{}{\mathbf{p}_o}$, $\prescript{m}{}{\mathbf{q}_{r}^c}$} $\forall \prescript{m}{}{\mathbf{q}_{r}} \in \mathcal{S}$

    $\mathcal{S} \leftarrow \mathcal{S} - \{ \prescript{m}{}{\mathbf{q}_{r}^*} \}$

    Navigation result: $r_{nav}$ $\leftarrow$ \KwNav{$\prescript{m}{}{\mathbf{q}_{r}^*}$}

    \lIf{$r_{nav} = false$}{\KwCont{}}

    Pickup result: $r_{pickup}$ $\leftarrow$ \KwPick{$\prescript{m}{}{\mathbf{p}_o}$}

    \KwRet{$r_{pickup}$}
}

% Rank $\mathcal{S}$ $\mapsto$ best candidate $\prescript{m}{}{\mathbf{q}_{r}^*}$

\caption{Object pickup based on planned robot placements.}
\label{alg:pickup-exec}
\end{algorithm}

\subsection{Picking up objects based on robot placement candidates}
\refalg{alg:pickup-exec} outlines how we carry out the object pickup using the robot placement candidates provided by \refalg{alg:base-placement-planning}.
The candidates are evaluated based on the combined distance from the robot's current position to the pose candidate and the distance from the pose candidate to the object, with the assumption that a longer distance would result in a more challenging grasp.
The resulting best candidate will be removed from $\mathcal{S}$ and used as the navigation goal for the robot to carry out the corresponding motions to pick up the object.
If navigation fails, the remaining candidates will be reevaluated based on the current robot pose, and a new best candidate will be chosen for executing the pickup behaviour.
This process is repeated until either the robot navigates successfully and begin the grasping motion, or if no candidate is left in $\mathcal{S}$.
Planning and execution of the navigation motion is done using a modified version of the \texttt{move\_base} ROS package\footnote{\url{https://wiki.ros.org/move_base}} provided by Toyota.
The subsequent manipulation motion is planned and executed using our existing solution previously described in~\cite{mitrevski2019noodle}.

\section{Experimental results and evaluation} \label{sec:eval}

To evaluate our approach, we carry out a set of grasping experiments using our robot placement planning algorithm and compare the results to a baseline grasp planning approach, in which the robot navigates to a predefined location and initiates the same object pickup behaviour.
Our experiments are carried out using a Human Support Robot (HSR)~\cite{yamamoto2019} from Toyota, which can be seen in \reffig{fig:experiment_setup} along with our experimental setup.
The remainder of this section will describe this setup and discuss the experimental results in detail.

\subsection{Experimental setup and procedure}

\begin{figure}[ht]
    \centering
    \includegraphics[width=0.6\textwidth]{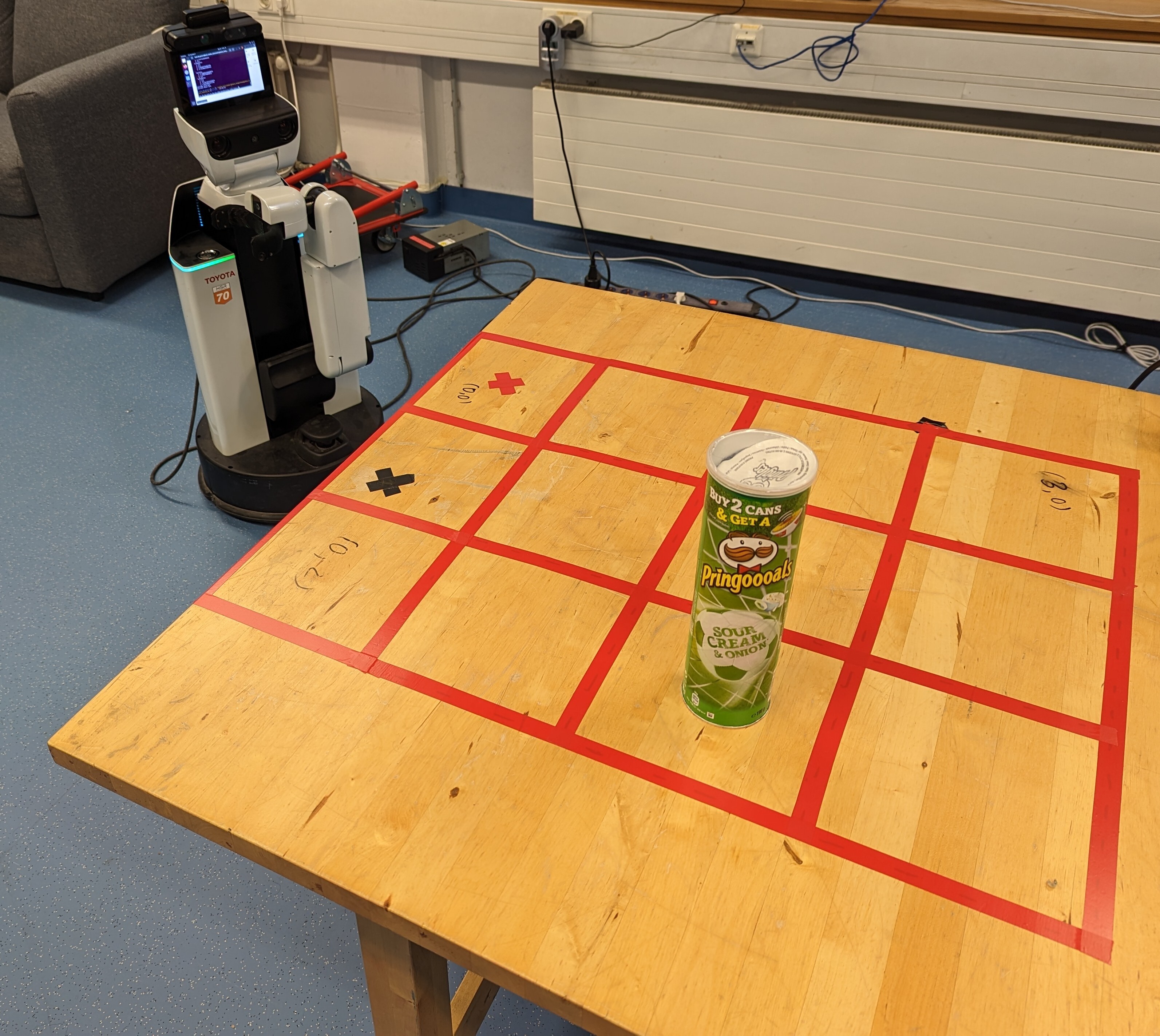}
    \caption{Experimental Setup}
    \label{fig:experiment_setup}
\end{figure}

In our experiments, the objects are placed on a standard dining table of dimensions (H $\times$ W $\times$ L) 74 $\times$ 80 $\times$ 180 cm.
To control for different object placements on the table surface, we tape a 3-by-4 grid on the table, whose cells are squares of dimensions 20 $\times$ 20 cm, as shown in~\reffig{fig:experiment_setup}.
Here, the long sides of the grid are 20 cm away from the long edges of the table, and one short side of the grid aligns with one short edge of the table.
We define a 2D discrete coordinate system for referring to specific grid cells, in which a cell can be denoted $(x, y)$ s.t. $x \in [0, 2]$ and $y \in [0, 3]$ are integers.
In~\reffig{fig:experiment_setup}, cell $(0, 0)$ is marked with a red ``x'', and the $x$-axis of the coordinate system aligns with the table's short edge.

For each approach, we perform five grasps of the same object for each grid cell, i.e. 120 grasp attempts in total.
A grasp is considered successful if the object remains in the gripper after the pickup behaviour, i.e. after the manipulator returns to its recovery configuration.
If the robot is unable to hold on to the object, or if collision with the table occurs, the attempt is considered a failure.
For the proposed robot placement planning approach, a human operator presents a ``Scan'' gesture to the robot to initiate object detection, then points to select the target object.
The robot then plans the robot placement poses and carry out the corresponding motions as described in \refsec{sec:approach}.
For the baseline approach, the robot detects and estimates the object's pose, navigates to a predefined location close to the short edge of the table, and initiates the pickup behaviour.

\subsection{Results and Discussion}

\begin{figure}[ht]
    \centering
    \includegraphics[width=0.8\textwidth]{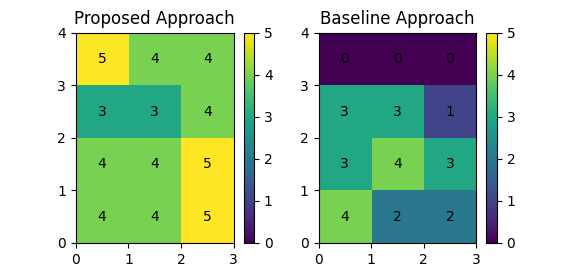}
    \caption{Heatmap of approach's success out of 5 trials in each grid.}
    \label{fig:heatmap}
\end{figure}

\begin{table}[ht]
\caption{Number of successful and failed attempts during the grasping experiments.}
\label{tab:grasp-results}

\centering
\begin{tabular}{X{0.1\linewidth}X{0.15\linewidth}X{0.15\linewidth}X{0.15\linewidth}X{0.15\linewidth}}
    \toprule
     \multirow{2}{*}{\textbf{Cell}} &  \multicolumn{2}{X{0.3\linewidth}}{\textbf{Proposed Approach}} & \multicolumn{2}{X{0.3\linewidth}}{\textbf{Baseline Approach}}
     \\ \cmidrule{2-5}
      & Success & Failure & Success & Failure
     \\ \midrule
     (0, 0) & 4 & 1 & 4 & 1 \\
     (1, 0) & 4 & 1 & 2 & 3 \\
     (2, 0) & 5 & 0 & 2 & 3 \\
     (0, 1) & 4 & 1 & 2 & 3 \\
     (1, 1) & 4 & 1 & 4 & 1 \\
     (2, 1) & 5 & 0 & 3 & 2 \\
     (0, 2) & 3 & 2 & 3 & 2 \\
     (1, 2) & 3 & 2 & 3 & 2 \\
     (2, 2) & 4 & 1 & 1 & 4 \\
     (0, 3) & 5 & 0 & 0 & 5 \\
     (1, 3) & 4 & 1 & 0 & 5 \\
     (2, 3) & 4 & 1 & 0 & 5 \\
     \bottomrule
\end{tabular}

\end{table}

The results of our grasping experiments are shown in \reftab{tab:grasp-results} and the heatmap of same is given in Fig.\ref{fig:heatmap}.
Overall, the proposed approach had a success rate of 81.7\% compared to 40\% of the baseline approach.
Attempts using the baseline approach are more likely to fail when the object is placed in the grids far from the short 
edge of the table, i.e. far from the predefined navigation goal.
When placed on the last 2 rows (cells $(x, 2)$ and $(x, 3)$), the object is near the limit of the robot's reach, resulting in the robot frequently colliding with the table.
% This is evident from the results presented in table \ref{tab:tabresults}, where grasping success significantly declined in the last two rows, with no successful grasps observed in the last row.
% Furthermore, the further the object, the more the robot arm had to extend to pick up the object; this reduced the balance of the robot, increasing the chances of it tipping over.
% Additionally, as this approach did not take the table into account, the robot frequently collided with the table trying to pick up objects in the last two rows, putting the robot's safety in jeopardy.
In contrast, the proposed approach allows the robot to navigate to other edges of the table before grasping, which makes picking up the object in the last row possible and even at a high success rate.
% The proposed approach shines in maximizing the grasp success of the object by utilizing the RGB-D and map data to account for the table and other objects between the robot and the target object along the path taken by the robot's arm to ensure that the robot will be able to pick up the target object once in position.
% This resulted in a base placement that maximized safety, as the base placements ensured that the robot didn't need to extend the arm as much as in the baseline approach, as the base was placed as close to the target object as possible while ensuring safety and success.
% This also helped in avoiding collisions with the table, unlike the baseline approach.
% One thing to note is that the robot found it a bit more difficult to pick up objects that were placed towards the centre of the table, which could be attributed to the physical limits of the robot and was a factor in the test with both approaches.

The experiments described thus far validate the soundness of the proposed robot placement planning algorithm and show how it enables a more flexible means to approach picking up objects, compared to the existing baseline approach.
However, they have not clearly shown the benefit of utilizing both RGB-D data and occupancy grid maps in the planning process.
This motivates additional experiments, e.g. grasping from a cluster of objects or with obstacles, to show whether the proposed approach can plan robot placements that reduce the risk of collision.

\section{Conclusion and Future Work} \label{sec:conclusion}

In this paper, we present an approach to finding a robot placement from where it can pick up an object.
Approaches addressing the same problem typically assume an end-effector pose is provided by some grasp planner, then find and evaluate robot placements based on this grasp pose.
Instead, we propose to directly find valid robot placement candidates using the estimated object position, RGB-D data and occupancy grid maps, and evaluate them using a simple metric.
The proposed approach is embedded in a Human-Robot Interaction pipeline, in which a human operator chooses the target object using hand gestures.
We evaluate our approach with a series of grasp experiments on the Toyota HSR, in which we compare the approach to an existing baseline implementation that sends the robot to a fixed location before initiating the pickup behaviour.
The results show that the proposed approach allows picking up objects from locations that are very challenging to the baseline implementation.

% Our proposed approach demonstrated a substantial success rate of 81.7\%, surpassing the baseline method's 40\%.
% Unlike the baseline approach, our method dynamically adjusts the robot's base placement based on object location, ensuring successful and safe object retrieval.
% Notably, it outperformed in grasping objects far from the table's short edge, addressing safety concerns, and minimizing arm extension.
% While facing some challenges with objects towards the table's centre, our approach showcased significant advantages in maximizing grasp success and avoiding collisions.
% Overall, the proposed approach proves effective in enhancing robotic manipulation tasks in diverse settings.

% In the future, we plan to implement active perception to dynamically adjust navigation and manipulation if required.
% We further want to use efficient techniques for point cloud processing, for instance, using voxel grids, to decrease the overall time taken by the algorithm.
As future work, several aspects of the proposed planning algorithm can be extended.
First, we used a simple distance range in our approach to represent the robot's reachability, but more elegant methods, such as reachability maps, can be employed that consider the robot's full kinematics.
Furthermore, while the cuboid between the object and a pose candidate serves as a simple approximation of the space occupied by the robot's arm while reaching for the object, it does not reflect the actual trajectory that the arm will follow in practice.
A more advanced approach may check for obstacles along the planned end-effector trajectory, e.g. to avoid collision in case the trajectory would go under the table surface.
Finally, instead of just combining the distance from the pose candidate to the object and the robot's current pose, better metrics can be employed to evaluate the candidates that take into account the potential navigation and manipulation motions that the robot needs to execute, such as one proposed by Yamazaki et al.~\cite{Yamazaki2022}.

\bibliographystyle{splncs04} % BibTeX users should specify bibliography style 'splncs04'.
\bibliography{bibliography.bib} % Use the bibliography.bib file as the source of references

\end{document}